\pdfoutput=1

\documentclass[11pt]{article}

\usepackage{EMNLP2022}

\usepackage{times}
\usepackage{latexsym}

\usepackage[T1]{fontenc}

\usepackage[utf8]{inputenc}

\usepackage{microtype}

\usepackage{inconsolata}

\usepackage{booktabs}
\usepackage{graphicx}

\title{End-to-End Evaluation of a Spoken Dialogue System for Learning~Basic~Mathematics}

\author{Eda Okur \\ Intel Labs, USA \\ \texttt{eda.okur@intel.com}
        \And Saurav Sahay \\ Intel Labs, USA \\ \texttt{saurav.sahay@intel.com}
        \AND Roddy Fuentes Alba \\ Intel Labs, Mexico \\ \texttt{roddy.fuentes.alba@intel.com}
        \And Lama Nachman \\ Intel Labs, USA \\ \texttt{lama.nachman@intel.com}
        \\}

\begin{document}
\maketitle
\begin{abstract}
The advances in language-based Artificial Intelligence (AI) technologies applied to build educational applications can present AI for social-good opportunities with a broader positive impact. Across many disciplines, enhancing the quality of mathematics education is crucial in building critical thinking and problem-solving skills at younger ages. Conversational AI systems have started maturing to a point where they could play a significant role in helping students learn fundamental math concepts. This work presents a task-oriented Spoken Dialogue System (SDS) built to support play-based learning of basic math concepts for early childhood education. The system has been evaluated via real-world deployments at school while the students are practicing early math concepts with multimodal interactions. We discuss our efforts to improve the SDS pipeline built for math learning, for which we explore utilizing MathBERT~\cite{MathBERT-2021} representations for potential enhancement to the Natural Language Understanding (NLU) module. We perform an end-to-end evaluation using real-world deployment outputs from the Automatic Speech Recognition (ASR), Intent Recognition, and Dialogue Manager (DM) components to understand how error propagation affects the overall performance in real-world scenarios.

\end{abstract}

\section{Introduction}

Following the advances in Artificial Intelligence (AI) research, building innovative applications to support education can present exciting opportunities with a positive and broader social impact. The United Nations (UN) Sustainable Development Goals\footnote{\url{https://sdgs.un.org/goals}} (SDGs)~\cite{desa2016transforming} represent an urgent call for action to help address critical global problems, where education is among the top five of these development areas (i.e., poverty, hunger, health, education, gender equality). Across many disciplines, improving the quality of mathematics education is crucial in building critical thinking and problem-solving skills at younger ages, which is a fundamental component of comprehensive and successful STEM education (i.e., science, technology, engineering, and mathematics). Language-based AI systems are starting to mature to a point where they could play a significant role in helping students learn and practice mathematical concepts. Despite its importance, applied Natural Language Processing (NLP) technologies for enhancing mathematics education still remain a highly under-explored area of research.

This study presents a task-oriented Spoken Dialogue System (SDS) developed to facilitate play-based learning of basic mathematical concepts for early childhood education. The system has been developed in the lab and evaluated via real-world deployments at school while the students are learning and practicing basic math concepts with multimodal interactions. These fundamental early math concepts and basic operations include constructing numbers using ones and tens, counting, addition, subtraction, measurement of length and size, etc. The multimodal interactions involve speech-based interactions to answer early math-related or game-related questions, counting and placing learning-specific tangible objects (i.e., manipulatives) in a visually observed space, touch-based interactions with the 1-to-100 number grid projected on the wall, to name a few.

This work discusses our efforts to improve the modular SDS pipeline built for game-based math learning and perform an end-to-end evaluation with various SDS components using real-world deployment outputs. The main SDS module we focus on investigating is Natural Language Understanding (NLU). The NLU arguably is the most critical component of goal-oriented dialogue systems that enables efficient communication between humans and intelligent conversational agents via application-specific comprehensive sub-tasks. Intent Recognition (IR) is at the heart of these NLU tasks, where the goal is to identify users' objectives from the input text and determine their intentions. 

The representation of human language is a crucial factor determining the success of conversational agents, especially in real-world applications. In the math learning domain, this language representation gains further significance due to the quite peculiar nature of mathematical language. With that motivation, we explored employing MathBERT~\cite{MathBERT-2021} representations for potential enhancement to the NLU module of our play-based math-learning system. MathBERT is a recently proposed language model created by pre-training the well-known BERT-base model~\cite{devlin-etal-2019-bert} on a large mathematical corpus. We compared the NLU results obtained by simply using BERT~\cite{devlin-etal-2019-bert} and ConveRT~\cite{ConveRT-2020} representations versus the new MathBERT representations. We further investigated the two variations of MathBERT models, one pre-trained with mathematics-specific vocabulary and the other with BERT-base vocabulary, to see their effects on our domain-specific math-learning NLU task.

As most of the application-specific and modular SDS pipelines do, our task-oriented SDS contains particular building blocks or modules for Automatic Speech Recognition (ASR), Natural Language Understanding (NLU), multimodal Dialogue Management (DM), Natural Language Generation (NLG), and Text-to-Speech (TTS). At its current stage of research \& development, for practical reasons, we employ template-based responses at the NLG module with off-the-shelf TTS. Thus, this study emphasizes more on speech recognition (ASR), intent understanding (NLU), and response selection (DM) tasks for a conversational agent that supports elementary math learning. For a complete end-to-end evaluation of our task-oriented SDS, we evaluated the ASR, NLU, and DM components to understand how error propagation affects the overall performance in real-world scenarios.

\section{Related Work}

\subsection{NLP for Mathematics Education}

Exploring the advancements in AI systems for social good and positive impact in the education domain, specifically to amplify students' learning experiences, has recently gained increasing interest from the research community~\cite{10.1145/2395123.2395128,CHASSIGNOL201816,MMHCI,baker2021artificial,zhai2021review}. Intelligent and interactive play-based learning systems have shown remarkable benefits for teaching mathematical concepts in smart and collaborative spaces~\cite{Lester_Ha_Lee_Mott_Rowe_Sabourin_2013,10.3389/feduc.2019.00081,10.1007/978-3-030-78292-4_28,sungamifying}. For early childhood education, a recent study by~\newcite{skene2022can} has showcased that game-based learning environments can offer significant advantages over traditional learning approaches while practicing fundamental math concepts, especially for younger learners. 

Harnessing NLP technologies to construct innovative applications for education is gaining popularity as an emerging area of research having various examples in the last decade~\cite{10.1007/978-3-319-19773-9_3,lende2016question,taghipour-ng-2016-neural,raamadhurai-etal-2019-curio,cahill-etal-2020-context,DBLP:journals/corr/abs-2112-01012,dutta-etal-2022-activity}. Within those efforts, exploring conversational agents for intelligent tutoring systems and smart education applications is a glaring sub-field of NLP in education, having several research studies tackling the problem from different angles~\cite{graesser2004autotutor,10.1007/978-1-84882-215-3_13,alex254848,PalaMohdNash2019gl,10.1145/3313831.3376781,DBLP:journals/corr/abs-2010-12710,zhang2021building}. 

Furthermore, there have been recent attempts to bridge the gap between general AI research and mathematics education~\cite{davila2017layout,jiang2018mathematics,mansouri2019tangent,yuan2020automatic,huangreal2,kumarphygital}. To further narrow our attention to language-based technologies applied to mathematics education, relatively few recent studies exist~\cite{shen2021classifying,suresh2022fine,loginova2022structural} which explore transfer learning to improve language representation for math-related tasks~\cite{DBLP:journals/corr/abs-2105-00377,MathBERT-2021}. Among these, MathBERT~\cite{MathBERT-2021} has been built specifically for challenging downstream NLP tasks in math education (e.g., knowledge component prediction, auto-grading open-ended question-answering, and knowledge tracing). It is indeed a mathematics-customized BERT model~\cite{devlin-etal-2019-bert}. MathBERT representations are created by pre-training the BERT-base model on large mathematical corpora, including pre-kindergarten, to high-school and college graduate-level mathematical text.


\subsection{Dialogue Systems and NLU}

For interactive early math learning applications, as we aim to build a spoken dialogue system for kids, we will briefly discuss the existing dialogue system technologies and language understanding approaches in a more generic context here.

Conversational agents or dialogue systems are mainly categorized as either open-ended or task-oriented~\cite{10.1145/3166054.3166058}. The open-ended dialogue systems or chatbots allow generic conversations such as chit-chat~\cite{serban2016building,Jurafsky:2000:SLP:555733}. On the other hand, task-oriented conversational AI systems are designed to accomplish specific tasks and handle goal-oriented conversations~\cite{serban2018survey, DBLP:journals/corr/abs-2009-13570}. With the advances of deep learning-based language technologies, improved access to high computing power, and increased availability of large datasets; the end-to-end trained dialogue systems can achieve encouraging results for both open-ended~\cite{serban2016building,DBLP:journals/corr/DodgeGZBCMSW15} and task-oriented~\cite{wen-etal-2017-network,DBLP:journals/corr/bordes2016learning,ham-etal-2020-end} applications. Dialogue Managers (DM) of task-oriented conversational AI systems are mostly sequential decision-makers. At that step, learning the optimal dialogue policies can be achieved via reinforcement learning (RL) from an excessive number of user interactions~\cite{zhao-eskenazi-2016-towards,shah2016interactive,cuayahuitl2017simpleds,dhingra-etal-2017-towards,liu2017e2e,su-etal-2017-sample}. However, building RL-based dialogue systems with highly limited user interaction data is immensely challenging. For this reason, supervised learning methods with traditional pipeline-based modular dialogue systems are still widely accepted when training data is initially limited to bootstrap the task-oriented SDS for further data collection~\cite{budzianowski-etal-2018-multiwoz}. For implicit dialogue context management, statistical and neural network-based dialogue system frameworks and toolkits~\cite{bocklisch2017rasa,ultes-etal-2017-pydial,burtsev-etal-2018-deeppavlov} are employed popularly in the research communities and industrial applications.

The NLU module of a dialogue system pipeline processes the user utterances as input text and usually predicts the user's intents or dialogue acts. For sequence learning tasks of Intent Recognition and Slot Filling~\cite{mesnil-2015,hakkani2016multi}, joint training of Intent Classification and Entity Recognition models have been explored widely~\cite{zhang-2016,Liu+2016,goo-etal-2018-slot,varghese2020bidirectional}. Many hierarchical multi-task learning architectures have been proposed for these joint NLU methods~\cite{h-zhou-2016,gu2017speech,wen-2018,AMIE-CICLing-2019,vanzo-etal-2019-hierarchical}. \newcite{DBLP:conf/nips/VaswaniSPUJGKP17} proposed the Transformer as a game-changing neural network architecture based entirely on attention mechanisms~\cite{DBLP:journals/corr/BahdanauCB14}. Right after the Transformers, Bidirectional Encoder Representations from Transformers (BERT)~\cite{devlin-etal-2019-bert} has been proposed. BERT has become one of the most significant breakthroughs in language representations research and has shown strong performance in several NLP tasks, including NLU. Lately,~\newcite{DIET-2020} introduced the Dual Intent and Entity Transformer (DIET) model as a lightweight multi-task architecture. DIET has been shown to outperform fine-tuning the BERT model for predicting intents and entities on a complex multi-domain NLU-Benchmark dataset~\cite{Liu2021}. For efficient language representation learning,~\newcite{ConveRT-2020} recently proposed the Conversational Representations from Transformers (ConveRT) model, which is another lightweight approach to obtain pre-trained embeddings as sentence representations successfully used in several conversational AI tasks.

\section{Language Understanding Methods}

For the NLU module within our early math-learning dialogue system pipeline, we have examined numerous options for Intent Recognition and built our NLU models on top of the Rasa open-source framework~\cite{bocklisch2017rasa}. 

\subsection{Baseline: TF+BERT}
The previous baseline Intent Recognition architecture available in the Rasa platform was based on supervised embeddings as part of the Rasa NLU~\cite{bocklisch2017rasa}. It was an embedding-based text classifier that embedded user utterances and intent labels into the same vector space. This former baseline architecture was inspired by the StarSpace algorithm~\cite{wu2018starspace}, where the supervised embeddings were trained by maximizing the similarity between intents and utterances. ~\newcite{KidSpace-SemDial-2019} enriched this embedding-based former baseline Rasa Intent Classifier by incorporating additional features and adapting alternative network architectures. To be more specific, they adapted the Transformer architecture~\cite{DBLP:conf/nips/VaswaniSPUJGKP17} and employed pre-trained BERT embeddings~\cite{devlin-etal-2019-bert} using the~\texttt{bert-base-uncased}~\footnote{\url{https://huggingface.co/bert-base-uncased}} model for Intent Recognition. We treat this simple and initial approach as our baseline NLU model in this study. We will refer to this method as TF+BERT in our experiments.

\subsection{DIET+BERT}
Next, we explored the potential benefits of adopting the recent DIET architecture~\cite{DIET-2020} for the Intent Recognition task in the basic math-learning domain. DIET is a transformer-based multi-task architecture for joint Intent Classification and Entity Recognition. The architecture includes a two-layer transformer shared for both NLU tasks. A sequence of entity labels is predicted with a Conditional Random Field (CRF)~\cite{DBLP:conf/icml/LaffertyMP01} tagging layer on top of the transformer output sequence corresponding to the input sentences treated as a sequence of tokens. For the intent labels, the transformer output for the classification token and the intent labels are embedded into the same semantic vector space. The dot-product loss is employed to maximize the similarity with the target label and minimize similarities with the negative samples. DIET architecture enables the incorporation of the pre-trained word and sentence embeddings from language models as dense features, with the flexibility to combine these with token-level one-hot and multi-hot encodings of character n-grams as sparse features. These sparse features are passed through a fully-connected layer with shared weights across all sequence steps. The output of the fully-connected layer is concatenated with the dense features from the pre-trained models. The high flexibility of this architecture allowed us to use any pre-trained embeddings as dense features in DIET, such as BERT~\cite{devlin-etal-2019-bert}, ConveRT~\cite{ConveRT-2020}, and MathBERT~\cite{MathBERT-2021}. To investigate the net benefits of DIET architecture, we adopted DIET with off-the-shelf pre-trained BERT embeddings using the~\texttt{bert-base-uncased} model~\cite{devlin-etal-2019-bert} and compared that against our baseline TF+BERT model. This approach (i.e., combining out-of-the-box BERT representations with the DIET classifier) will be referred to as DIET+BERT in the experiments\footnote{Refer to~\newcite{DIET-2020} for hyper-parameters, computational costs, and hardware specifications.}.

\subsection{DIET+ConveRT}
Conversational Representations from Transformers (ConveRT)~\cite{ConveRT-2020} is a promising architecture recently proposed to learn pre-trained representations that are well-suited for conversational AI applications, especially for the real-world Intent Classification tasks. ConveRT is a transformer-based dual-encoder network leveraging quantization and sub-word level parameterization, where the pre-trained representation outputs from its sentence encoder can be utilized for the conversational Intent Classification tasks. DIET and ConveRT are both lightweight architectures with faster and more efficient training capabilities than their counterparts. When incorporating the ConveRT embeddings within the DIET classifier, the initial embeddings for the classification tokens are set as the input sentence encoding obtained from the ConveRT model. That enables exploiting extra contextual information from the complete sentence on top of the word embeddings. For these reasons, we adopted the DIET architecture and utilized pre-trained ConveRT embeddings to potentially improve the Intent Recognition performances on our domain-specific early math-learning datasets. We will call this approach DIET+ConveRT in our experiments.

\subsection{DIET+MathBERT}
Finally, a pre-trained language model named MathBERT~\cite{MathBERT-2021} has been presented lately for downstream NLP tasks in the domain of mathematics education. MathBERT is a BERT-like language representation model further pre-trained from the~\texttt{bert-base-uncased} model with dedicated mathematical corpora. Note that BERT is a general-purpose language model trained on a vast amount of unlabeled text corpus (i.e., Wikipedia and BookCorpus) with 3.3 billion words, which can be further pre-trained to obtain a new set of model weights for transfer learning. On the contrary, MathBERT is pre-trained on 100 million tokens of mathematical corpora, including instructional texts from books, curriculum, Massive Open Online Courses (MOOCs), and arXiv.org paper abstracts, covering all possible grade levels from pre-k to college graduate-level content. Although the scale of training corpora is much smaller than the BERT-base model, MathBERT can still have the potential to be more effective in math-related NLP tasks. That is hypothesized because mathematical language frequently uses domain-specific vocabulary and concepts that require better word representations within the math context. With that motivation, we utilized the publicly available MathBERT release~\footnote{\url{https://github.com/tbs17/MathBERT}}, which includes the PyTorch and TensorFlow versions of MathBERT models and the tokenizers. We explored adopting the DIET architecture with pre-trained MathBERT embeddings to empower the NLU/Intent Recognition task in our dialogue system designed for teaching basic math concepts. Furthermore, we investigated the representations from the MathBERT-base model that is trained with the BERT-base vocabulary (i.e., origVocab) and compared that against the representations from MathBERT-custom model pre-trained with math-specific vocabulary (i.e., mathVocab). This custom mathVocab set is also released to reflect the specific nature of mathematical jargon and concepts used in math corpora. In our NLU experiments, we will refer to these two distinct approaches as DIET+MathBERT-base (origVocab) and DIET+MathBERT-custom (mathVocab).

\begin{table}[!t]
  \centering
  \resizebox{\columnwidth}{!}{
  \begin{tabular}{lcc}
    \toprule
     & \multicolumn{2}{c}{\textbf{Math-Game Data}} \\
    \textbf{Statistics} & Planting & Watering \\
    \midrule
    \# Distinct Intents & 14 & 13 \\
    Total \# Samples (Utterances) & 1927 & 2115 \\
    \# Math-related Samples & 452 & 599 \\
    Min \# Samples per Intent & 22 & 25 \\
    Max \# Samples per Intent & 555 & 601 \\
    Avg \# Samples per Intent & 137.6 & 162.7 \\
    Min \# Words per Sample & 1 & 1 \\
    Max \# Words per Sample & 74 & 65 \\
    Avg \# Words per Sample & 5.26 & 4.95 \\
    \# Unique Words (Vocab) & 1314 & 1267 \\
    Total \# Words & 10141 & 10469 \\
    \bottomrule
  \end{tabular}
  }
  \caption{KidSpace-POC NLU Dataset Statistics}
  \label{data-st-poc}
\end{table}

\section{Experimental Results}

\subsection{Math-Game Datasets}

Our experiments are conducted on the NLU datasets of Kid Space Planting and Watering use cases~\cite{KidSpace-ICMI-2018,KidSpace-ETRD-2022}, having utterances from gamified math learning experiences designed for early childhood education (i.e., 5-to-8 years old kids). The intelligent conversational agent should accurately understand these children's utterances and provide appropriate feedback. The use cases include a specific flow of interactive games facilitating elementary math learning. The FlowerPot (i.e., Planting) game builds on the math concepts of tens and ones, with the larger flower pots representing tens and smaller pots representing ones. The virtual character provides the number of flowers the children should plant, and when the children have placed the correct number of large and small pots against the wall, digital flowers appear. In the NumberGrid (i.e., Watering) game, the children are presented with basic math questions with clues. When the correct number is touched on the 1-to-100 number grid projected on the wall, water is virtually poured to water the flowers. The virtual character supports the kids with learning to construct numbers using the `tens' and `ones' digits, practicing simple counting, addition, and subtraction operations. These math-game datasets have a limited number of user utterances, which are annotated manually for intent types defined for each learning activity. Some of the intents are highly generic across learning activities (e.g., \textit{affirm}, \textit{deny}, \textit{next-step}, \textit{out-of-scope}, \textit{goodbye}), whereas others are highly domain-dependent and game-specific (e.g., \textit{intro-meadow}, \textit{answer-flowers}, \textit{answer-water}, \textit{answer-valid}, \textit{answer-invalid}) or math-learning/task-specific (e.g., \textit{ask-number}, \textit{counting}).

\begin{table}[!t]
  \centering
  \resizebox{\columnwidth}{!}{
  \begin{tabular}{lcc}
    \toprule
     & \multicolumn{2}{c}{\textbf{Math-Game Data}} \\
    \textbf{Statistics} & Planting & Watering \\
    \midrule
    \# Distinct Intents & 12 & 11 \\
    Total \# Samples (Utterances) & 2173 & 2122 \\
    \# Math-related Samples & 549 & 602 \\
    Min \# Samples per Intent & 4 & 6 \\
    Max \# Samples per Intent & 1005 & 1005 \\
    Avg \# Samples per Intent & 181.1 & 192.9 \\
    Min \# Words per Sample & 1 & 1 \\
    Max \# Words per Sample & 45 & 44 \\
    Avg \# Words per Sample & 4.80 & 4.48 \\
    \# Unique Words (Vocab) & 772 & 743 \\
    Total \# Words & 10433 & 9508 \\
    \bottomrule
  \end{tabular}
  }
  \caption{KidSpace-Deployment NLU Dataset Statistics}
  \label{data-st-dep}
\end{table}

The NLU models are trained and validated on the initial proof-of-concept (POC) datasets~\cite{KS2.0-NAACL-2021} to bootstrap the agents for real-world deployments. These POC datasets were curated manually to train the initial SDS models based on User Experience (UX) studies. The models are then validated on the UX sessions in the lab with five kids going through these early math-learning games. Table~\ref{data-st-poc} shows the statistics of these KidSpace-POC NLU datasets. Planting and Watering game POC datasets have 1927 and 2115 user utterances, respectively. The deployment datasets were collected later from twelve kids, where the math-learning system was deployed in a classroom at school~\cite{okur-etal-2022-nlu}. Table~\ref{data-st-dep} shows the statistics of KidSpace-Deployment NLU datasets, where Planting and Watering deployment datasets have 2173 and 2122 user utterances, respectively. These deployment datasets are used only for testing purposes, where we train our NLU models on the POC datasets. For both in-the-lab and real-world datasets, the spoken user utterances and agent responses are transcribed manually at first. These transcriptions are annotated for the user intent and agent response types we defined for each math learning activity. These transcribed and annotated final utterances are analyzed and used in our experiments.

\begin{table}[!t]
  \centering
  \resizebox{\columnwidth}{!}{
  \begin{tabular}{lcc}
    \toprule
     & \multicolumn{2}{c}{\textbf{Math-Game Data}} \\
    \textbf{Model} & Planting & Watering \\
    \midrule
    TF+BERT (baseline) & 90.50$\pm$0.25 & 92.43$\pm$0.32 \\
    \midrule
    DIET+BERT & 94.00$\pm$0.38 & 96.39$\pm$0.14 \\
    DIET+ConveRT & \textbf{95.88}$\pm$\textbf{0.42} & \textbf{97.69}$\pm$\textbf{0.11} \\
    \midrule
    DIET+MathBERT-base & 90.40$\pm$0.16 & 93.56$\pm$0.26 \\
    DIET+MathBERT-custom & 90.82$\pm$0.10 & 93.67$\pm$0.10 \\
    \bottomrule
  \end{tabular}
  }
  \caption{NLU/Intent Recognition micro-avg F1-scores (\%): TF+BERT (baseline), DIET+BERT, DIET+ConveRT, DIET+MathBERT-base (origVocab), and DIET+MathBERT-custom (mathVocab) models trained and validated on KidSpace-POC datasets.}
  \label{nlu-results-poc}
\end{table}

\subsection{DIET Classifier with MathBERT Representations for NLU}

Tables~\ref{nlu-results-poc} and~\ref{nlu-results-dep} present a summary of our NLU/Intent Recognition experimental results for the Planting and Watering math-game use cases at school, covering a series of task-oriented interactions for early math education. On the proof-of-concept (POC) datasets that we created for each math-game, we achieved above 95\% F1-scores for Intent Recognition performances with our best-performing DIET+ConveRT NLU models (see Table~\ref{nlu-results-poc}). We trained and cross-validated these NLU models on math-game or activity-specific datasets having around 2K POC samples. When we later tested these models on real-world deployment data collected at school, we observed F1-score performance drops of around 7\% with our best-performing DIET+ConveRT models (see Table~\ref{nlu-results-dep}). This performance drop is anticipated and explainable as we operate on the noisier real-world utterances collected from younger children in dynamic play-based environments.

\begin{table}[!t]
  \centering
  \resizebox{\columnwidth}{!}{
  \begin{tabular}{lcc}
    \toprule
     & \multicolumn{2}{c}{\textbf{Math-Game Data}} \\
    \textbf{Model} & Planting & Watering \\
    \midrule
    TF+BERT (baseline) & 85.08$\pm$0.49 & 90.06$\pm$0.56 \\
    \midrule
    DIET+BERT & 87.03$\pm$0.30 & 89.63$\pm$0.62 \\
    DIET+ConveRT & \textbf{89.00}$\pm$\textbf{0.29} & \textbf{90.57}$\pm$\textbf{0.86} \\
    \midrule
    DIET+MathBERT-base & 84.52$\pm$1.19 & 86.85$\pm$1.15 \\
    DIET+MathBERT-custom & 85.22$\pm$0.78 & 87.80$\pm$1.45 \\
    \bottomrule
  \end{tabular}
  }
  \caption{NLU/Intent Recognition micro-avg F1-scores (\%): TF+BERT (baseline), DIET+BERT, DIET+ConveRT, DIET+MathBERT-base (origVocab), DIET+MathBERT-custom (mathVocab) models trained on KidSpace-POC \& tested on KidSpace-Deployment datasets.}
  \label{nlu-results-dep}
\end{table}

\begin{table*}[t!]
  \centering
  \footnotesize
  \begin{tabular}{ccccc}
    \toprule
      & \textbf{Math-Game} & \textbf{\# Test Samples} & \textbf{NLU/Intent Recognition} & \textbf{ASR+NLU} \\
    \textbf{NLU Model} & \textbf{Activity} & \textbf{(Speech/Utterances)} & \textbf{F1 (\%)} & \textbf{F1 (\%)} \\
    \midrule
    DIET+ConveRT & Planting & 588 & 91.8 & 75.8 \\
    \midrule
    DIET+ConveRT & Watering & 664 & 97.4 & 84.7 \\
    \bottomrule
  \end{tabular}
  \caption{End-to-End (ASR + NLU/Intent Recognition) Evaluation Results on Planting and Watering Math-game activity datasets: DIET+ConveRT models trained on KidSpace-POC datasets and tested on KidSpace-Deployment datasets.}
  \label{e2e_asr_nlu}
\end{table*}

\begin{table*}[t!]
  \centering
  \footnotesize
  \resizebox{\textwidth}{!}{
  \begin{tabular}{cccccccc}
    \toprule
    \textbf{Math-Game} & \textbf{\#test-NLU} & \textbf{NLU/Intent} & \textbf{ASR+NLU} & \textbf{\#test-DM} & \textbf{DM/Response} & \textbf{NLU+DM} & \textbf{ASR+NLU+DM} \\
    \textbf{Activity} & \textbf{(utterances)} & \textbf{F1 (\%)} & \textbf{F1 (\%)} & \textbf{(responses)} & \textbf{F1} & \textbf{F1} & \textbf{F1} \\
    \midrule
    Planting & 184 & 90.5 & 73.3 & 209 & 0.89 & 0.87 & 0.82 \\
    \midrule
    Watering & 346 & 95.2 & 84.4 & 403 & 0.93 & 0.91 & 0.89 \\
    \bottomrule
  \end{tabular}
  }
  \caption{End-to-End (ASR + NLU/Intent Recognition + DM/Response Selection) Evaluation Results on Planting and Watering Math-game activity datasets: DIET+ConveRT NLU models and TED DM models trained on KidSpace-POC datasets and tested on KidSpace-Deployment datasets.}
  \label{e2e_asr_nlu_dm}
\end{table*}

To the best of our knowledge, this study presents the first attempt to adopt the lightweight multi-task DIET architecture and incorporate pre-trained MathBERT embeddings as dense features. We combine these MathBERT representations with sparse word and character-level n-gram features in a plug-and-play fashion. The motivation behind this was empowering math domain-specific embeddings for the NLU task targeted at kids playing basic math-learning games. However, we could not observe any benefits of employing MathBERT-base (i.e., pre-trained using origVocab of BERT-base) or MathBERT-custom (i.e., pre-trained using customized mathVocab) representations for the Intent Recognition task. Although MathBERT-custom seems to perform slightly better than MathBERT-base as expected, the gain is insignificant and still way lower than ConveRT and even BERT. There could be many reasons for this, such as the possible mismatch between our domain and advanced mathematical corpora (e.g., mathematical equations and symbols) with graduate-level textbooks that MathBERT trained on. Compared to this, our early childhood math education domain involves more basic concepts and simple operations (e.g., ones and tens, counting, adding, and subtracting). In addition, MathBERT is pre-trained on a relatively small set (i.e., 100M tokens) compared to the massive general-purpose corpora that the BERT models are trained on (3.3B words). ConveRT embeddings have already been shown to perform well on conversational tasks such as Intent Classification, partly because these are pre-trained on large corpora of natural conversational datasets (e.g., Reddit conversational threads). The success of ConveRT representations over MathBERT could also indicate that our educational game datasets involve numerous utterances around play-based conversations tailored towards planting and watering flower use cases. Compared to those, our datasets include limited interactions that directly involve numbers or counting/addition/subtraction operations. Our intent class distributions also support this observation, where we have around 450-600 samples within approximately 2K utterances in POC datasets annotated with directly math-related intents (e.g., \textit{ask-number}, \textit{counting}). In the deployment datasets, we observed around 550-600 math-related utterances within a total of 2.1K samples. Nevertheless, instead of pre-training the BERT models to create MathBERT, one can also explore pre-training the ConveRT model on large and more elementary math-related corpora as the next step.

\begin{table*}[t!]
  \centering
  \footnotesize
  \begin{tabular}{cllcc}
    \toprule
    \textbf{Math-Game} & \textbf{Human Transcription} & \textbf{ASR Output} & \textbf{Intent} & \textbf{Prediction} \\
    \midrule
    Planting & Sunflowers & 7 flour & \textit{answer-flowers} & \textit{counting} \\
    & We need just one more & jasmine ranvir & \textit{counting} & \textit{answer-flowers} \\
    & Twenty four & trailer for & \textit{counting} & \textit{intro-game} \\
    & We just counted nineteen & he doesn't canton my team & \textit{counting} & \textit{out-of-scope} \\
    & Twenty two & tell me too & \textit{counting} & \textit{out-of-scope} \\
    & Six & snakes & \textit{counting} & \textit{answer-valid} \\
    & Twelve & towel & \textit{counting} & \textit{answer-invalid} \\
    & Nine & nah & \textit{counting} & \textit{deny} \\
    \midrule
    Watering & Water could help them bloom
 & why don't can count them you & \textit{answer-water} & \textit{counting} \\
     & Thirteen flowers & sure thing flowers & \textit{counting} & \textit{affirm} \\
    & Seven & I haven't & \textit{counting} & \textit{deny} \\
    & Twenty eight & try it & \textit{counting} & \textit{out-of-scope} \\
    & Three & tree & \textit{counting} & \textit{answer-valid} \\
    & Five & bye & \textit{counting} & \textit{goodbye} \\
    & Bye bye Oscar	& buy a hamster & \textit{goodbye} & \textit{answer-invalid} \\
    \bottomrule
  \end{tabular}
  \caption{ASR + NLU/Intent Recognition Error Samples from Kid Space Planting and Watering Math-game datasets.}
  \label{errors}
\end{table*}

\subsection{End-to-End Evaluation}

Our SDS pipeline starts by recognizing user speech via the ASR module and feeds the recognized text into our NLU component. We developed NLU models performing Intent Recognition to interpret user utterances. Then we pass these user intents together with multimodal inputs, such as user actions and objects, into the DM component. The multimodal dialogue manager handles verbal and non-verbal communication inputs from the NLU (e.g., intents) and external nodes processing audio-visual information (e.g., poses, gestures, objects, game events, and actions). We pass these multimodal inputs directly to the DM in the form of relevant multimodal intents. The Dialogue State Tracking (DST) model tracks what has happened (i.e., the dialogue state) within the DM. Then, the output of DST is used by the Dialogue Policy to decide which action the system should take next. Our DM models predict the appropriate agent actions and responses based on all the available contextual information (i.e., language-audio-visual inputs, game events, and dialogue history/context from previous turns). When the DM predicts verbal response types, the NLG module retrieves actual bot responses that are template-based in our use cases. We create a variety of response text by preparing multiple templates for each response type, where the final response template is randomly assigned at run-time. Finally, the generated text responses are sent to the TTS module to output agent utterances. Please refer to~\newcite{okur-etal-2022-data} for our multimodal SDS pipeline diagram. 

We were assuming perfect (or human-level) speech recognition performances for the NLU results obtained on manual transcriptions (by human transcribers) until now. However, we observed around 30\% word-error-rate (WER) in ASR transcriptions for kids' speech. These ASR outputs are obtained via the top hypothesis given by Google Cloud Speech-to-Text API~\footnote{\url{https://cloud.google.com/speech-to-text/}}. Considering these ASR errors propagating into the ASR+NLU pipeline, the Intent Recognition F1-score performances drop around 11-to-17\% (see Tables~\ref{e2e_asr_nlu} and~\ref{e2e_asr_nlu_dm}) when evaluated directly on noisy ASR outputs acquired using Google ASR engine. We are currently working towards improving the Automatic Speech + Intent Recognition (i.e., ASR+NLU) performances by exploring the N-best ASR hypotheses~\cite{ganesan-etal-2021-n} instead of using only the top ASR hypothesis.

For the DM model development, we adopted a recently proposed Transformer Embedding Dialogue (TED) policy architecture~\cite{TED-2019}, which is highly suitable to our multimodal math-learning use cases. In TED architecture, a transformer's self-attention mechanism operates over the sequence of dialogue turns to select the appropriate agent response. Despite the noisy ASR outputs with relatively higher WER in kids' speech compared to adults, when we performed end-to-end evaluations with the ASR+NLU+DM pipeline, we observed only 4-to-7\% drops in response prediction F1-score performances (see Table~\ref{e2e_asr_nlu_dm}). That means error propagation from ASR and NLU has much less effect on the Dialogue Manager (DM) outputs, which are the agent's final actions and selected responses.

\subsection{Error Analysis}

Table~\ref{errors} presents several concrete examples to compare the utterance text obtained by manual human transcriptions (i.e., ground truth) versus problematic ASR outputs (i.e., speech transcriptions). The ground truth intent classes based on gold data annotations on human transcriptions are shown along with the predicted intent classes on ASR outputs obtained by our best-performing DIET+ConveRT NLU models. These ASR errors, especially on domain-specific math-related intents, could explain how errors propagate into the NLU module of our SDS pipeline and significantly degrade the performance of the Intent Recognition task. Although such ASR errors are expected in noisy real-world application data, especially with kids of age 5-to-8~\cite{dutta-etal-2022-activity}, this analysis also points to a significant room for improvements in the ASR+NLU pipeline. It also encourages us to explore mitigation strategies such as utilizing N-best ASR outputs~\cite{ganesan-etal-2021-n} and employing phonetic-aware representations~\cite{sundararaman2021phoneme} that can be more robust to ASR errors.

\section{Conclusion and Future Work}

Improving the quality of mathematics education is vital in developing problem-solving and critical thinking skills for younger learners, which are fundamental for comprehensive STEM education. This study showcased a task-oriented SDS built to promote play-based math learning for early childhood education. The conversational AI system is implemented and evaluated on real-world deployment data collected in classrooms while the kids are practicing basic math concepts. We presented our attempts to enrich the modular SDS pipeline for gamified math learning and prosecuted an end-to-end evaluation using several SDS components tested on real-world deployment data. For NLP applied to math education, language representations can play a significant role due to the exceptional nature of math language. We investigated employing the language representations from the MathBERT model created by pre-training the BERT-base on mathematical corpora. We compared the Intent Recognition results obtained using BERT and ConveRT representations versus the recently proposed MathBERT embeddings on top of the DIET architecture for NLU. To perform an end-to-end evaluation of our SDS pipeline, we evaluated the ASR, NLU, and DM modules to investigate how error propagation affects the overall SDS performance in real-world math-learning scenarios.

In future work, we aim to explore adopting the N-Best-ASR-Transformer architecture~\cite{ganesan-etal-2021-n} to utilize multiple ASR hypotheses. This approach can improve the Intent Recognition performances and mitigate recognition errors propagated into the ASR+NLU+DM pipeline due to using only the top ASR hypothesis. Another future direction is to enhance math-specific language representation learning by pre-training the ConveRT model on large math corpora, especially tailored towards early math education (e.g., pre-k to 2nd-grade math curriculum), and then fine-tuning them on our NLU tasks for game-based math learning.


\section*{Limitations}



Before discussing the limitations of our study, note that the goal of this multimodal dialogue system that we built is to improve the quality of mathematics education for younger learners. To begin with, the cost of the overall setup currently deployed at school (e.g., projector, RGB-D/3D cameras, LiDAR sensor, lapel microphones) can be a limitation, especially for public schools in disadvantaged regions. That can potentially prevent us from having a broader positive impact with our AI for social-good efforts.

Another limitation of this work is the size of the collected datasets. Since the multimodal data collection from authentic classrooms and their labor-intensive annotation process is costly, we need to be innovative to work on such low-data regimes and benefit from the transfer learning paradigm whenever possible. Unfortunately, this data scarcity also affects the generalizability and reliability of our experimental results and end-to-end evaluations, which affects the overall robustness of such real-world applications.

In addition to the deployment costs and data-size concerns, we are bound to use lapel microphones to capture the speech from subjects. That affects the overall unobtrusiveness of the system. Although our ultimate goal is to use the microphone arrays in the classroom, the high WER observed in ASR with kids' speech, even with lapel mics, prevents us from using these far-field mic-array recordings.

Finally, instead of pre-training the BERT model on large math corpora (as performed to create MathBERT), we aimed to pre-train the ConveRT model on a more early-math-related subset of the corresponding corpora. However, although the code and models are publicly available, the authors of MathBERT~\cite{MathBERT-2021} have not released the fine-tuning math dataset per the data owner's request. Hence, we cannot perform these DIET+MathConveRT experiments before we collect our early-math-related corpora for transfer learning, which is another limitation of this work.

\section*{Ethics Statement}
It is worth noting that before the proof-of-concept research deployments, a rigorous Privacy Impact Assessment process has been followed. Legal approvals have been sought and received to conduct research with parents, their children, and educators. All research participants signed consent forms before the studies, informing them of all vital details about the studies, including the research goals and procedures, along with how their data would be collected and used to support our research. As with all research projects, our collaborators abide by strict data privacy policies and adhere to ongoing oversight.

\section*{Acknowledgements}

We gratefully acknowledge our current and former colleagues from the Intel Labs Kid Space team, especially Ankur Agrawal, Glen Anderson, Sinem Aslan, Benjamin Bair, Arturo Bringas Garcia, Rebecca Chierichetti, Hector Cordourier Maruri, Pete Denman, Lenitra Durham, David Gonzalez Aguirre, Sai Prasad, Giuseppe Raffa, Sangita Sharma, and John Sherry, for the conceptualization and the design of school use-cases to support this research. We also would like to show our gratitude to the Rasa team for the open-source framework and the community developers for their contributions that enabled us to conduct our research and evaluate proof-of-concept models for our usages.

\bibliography{anthology,custom}
\bibliographystyle{acl_natbib}

\end{document}